# Automatic Parallel Corpus Creation for Hindi-English News Translation Task

1st Aditya Kumar Pathak
*dept. Computer Science and Engineering*
IIIT Bhubaneswar
Bhubaneswar, India
adityapathak.cse@gmail.com

2nd Priyankit Acharya
*dept. Computer Science and Engineering*
IIIT Bhubaneswar
Bhubaneswar, India
priyankit.97@gmail.com

3rd Dilpreet Kaur
*dept. Computer Science and Engineering*
IIIT Bhubaneswar
Bhubaneswar, India
a116008@iiit-bh.ac.in

4th Rakesh Chandra Balabantaray
*dept. Computer Science and Engineering*
IIIT Bhubaneswar
Bhubaneswar, India
rakesh@iiit-bh.ac.in

*Abstract*— **Parallel corpus for multilingual NLP tasks, deep learning applications like Statistical Machine Translation Systems is very important. The parallel corpus of Hindi-English language pair available for news translation task till date is of very limited size as per the requirement of the systems are concerned. In this work we have developed an automatic parallel corpus generation system prototype, which creates Hindi-English parallel corpus for news translation task. Further to verify the quality of generated parallel corpus we have experimented by taking various performance metrics and the results are quiet interesting.**

I. INTRODUCTION

Machine translation through data driven approach is one of the most important research topic in the field of MT. Basically a parallel corpus i.e. the collection of large texts of different languages aligned in a parallel manner or are translation of each other, act as data to be trained for both statistical machine translation system as well as Neural machine translation system. As news articles are the part of our day to day life, we need a news translation task to generate news in different languages. For any translation task parallel corpus acts as an indispensable resource. At present the parallel corpus available for news translation is very limited, which is proving to be a barrier for doing an extensive research in this domain.

Given any translation task, accuracy of the system acts as a main factor to determine its final implications. The accuracy of the translation task between language pairs can be improved with the help of availability of large parallel corpus. Creating a large parallel corpus manually is a very tedious, time consuming and expensive task. As a consequence, in most of the translation tasks, comparable corpora are used as a resource which accordingly affects the performance and accuracy of the system.

In our work, we have developed a prototype for automatic generation of Hindi-English news parallel corpus. The parallel corpus was made from comparable corpus crawled from the web from various sources, with an improved quality based on fuzzy string matching algorithm. Further the quality of parallel corpus was analyzed by taking various performance metrics into consideration.

II. LITERATURE SURVEY

The main issue to be addressed in a Machine Translation system, is the creation of parallel corpus. There are many parallel corpus available for resource rich languages but, from a practical perspective, creation of small parallel corpora manually could be relatively simpler and easier, whereas, creation of large one manually is hard as well as time consuming. Other than this, maintaining the translation quality between the language pairs act as a core part for building a corpus which further aids to the performance of an MT system between the respective language pair. Here are some of the issues which are to be taken care of, while creating a parallel corpus:

(i) Optimizing parallel corpus to carry out translation between two language pairs.

(ii) The essential parameters required for creating a parallel corpus.

(iii) The efficient size of parallel corpora for more accurate translation result.

Other than all these issues, we need to find out what kind of parallel or comparable corpora will be more suitable for our translation task.

Gale and Church [1] used an approach based on number of characters present within the sentences. The idea used by mentioned that, long sentences would be translated into long ones and the short sentences with the short ones. This method worked aptly on French to English language pair but, the performance went down with the languages which have low length correlation such as Chinese to English. This method also failed for other language pairs having high length correlation. Other researchers worked with methods combining sentence length statistics with vocabulary alignment.

P. Sheridan (et.al, 1996) [2] gave an approach based on query expansion method of thesaurus for multilingual information retrieval. This was done with set of multilingual documents

Previous attempts for sentence alignment for parallel corpus creation were generally based on vocabulary alignment combined with sentences length. Brown's method [3] worked by finding out sentence length with the number of words.

Parallel as well as comparable corpora both are used as a resource for translation process as well as for comparative studies (McEnery and Xiao, 2007)[4], when used in association with parallel corpora, comparable corpora act as a useful resource for comparative study.

Sunita Arora (et.al. 2010) [5] discussed on two pass approach at sentence level to do automatic alignment, for automatic creation of Hindi-Punjabi parallel corpus with the help of comparable corpora.

Researchers continued to explore methods based on combination for alignment of vocabularies with that of statistics for sentence lengths. [6, 7]. Text aligners with these implemented processes are available such as BLEU align, that is based on the project developed by the University of Zurich. In supplement to parallel text, one text alignment is required by BLEU align. Bilingual Evaluation Understudy or BLEU similarity measure is used by it which is based on length for aligning of texts [8]. One of the open source tool also includes Hunalign tool [9], developed by the Media Research Center. As in the method suggested by Gale and Church, sentence lengths as well as a dictionary for aligning of text contained within two languages on a sentence level. The issue of ordering of sentence is not addressed in the method used by Hunalign. A commercial product is made by ABBYY Group called ABBYY aligner. Proprietary word databases are used by this product for aligning of text portions within the sentences [10]. There is another open source product called the Unitex Aligner [11]. XAlign tool [12], is being used by it which itself is based on the length of characters contained within the paragraph or sentences [13].

As, the comparable corpora can be proven bad for comparative studies if resultant sampled frames are not fully comparable. The comparable corpora with the qualities of non-parallel corpora has an ability to undertake these limitations of parallel corpora and best takes care of the two limitations of parallel corpora, as sources for original monolingual texts are much more in abundance as compared to translated texts. Though, as compared with parallel corpora, mining translations in comparable corpora becomes a more challenging task. One big problem is that many of the texts that are parallel in the comparable corpora, though they contain the parallel fragments, contain non-parallel fragments as well, mostly at the beginning or at the end.

The parallel fragments can be noticed anywhere in the document pair. Beginning and ending of parallel fragments can be found anywhere within the text with the possibility of skipping one or more than one number of sentences without the breakage of fragment. For the extraction of terminologies, specialized comparable and parallel corpora are clearly used, while for the comparison of general linguistic features like tense and aspect, balanced corpora are presumed to be more representative of any given language in general. Specialized parallel corpora can be useful in domain-specific translation research. The comparable corpora must contain sampling of frames but, in parallel corpora sampling of frames are not at all required, as each corpus component is the exact translation of the other. In most of the comparable corpora finding comparable text types is easier in different languages. Therefore, when compared with parallel corpora, it is easier for comparable corpora to be designed as a general balanced corpora. Parallel corpora acts as a unique resource for MT system development and are much used for providing assistance to human translators. Parallel corpora have been used to develop CAT or Computer Assisted Translation tools for human translators, such as Translation Memories (TM), bilingual concordances and translator oriented word processors systems.

By looking forward to the advantages as well as the problems that were present in the parallel corpora till now, we are trying to make a more accurate parallel corpora with large number of sentences.

### III. RESOURCES

We have crawled Hindi news of different genres from the Navbharat Times and for comparison purpose, we have crawled English news from many different web pages like the Times of India, the Hindu, Quora and many other news sources.

### IV. METRICS FOR CHECKING THE QUALITY OF PARALLEL CORPUS

The description of the performance metrics selected in order to describe the quality of our parallel corpus are defined in following sections:

#### a. Gestalt pattern matching

The idea of Gestalt pattern matching [14] algorithm is to discover longest adjacent identical subsequence which includes no 'junk' elements. This subsequence is termed as an anchor. Similar approach is then further implemented recursively to the fractions of subsequences towards the left and right of the anchor.

#### b. Hamming distance

Hamming distance is one of the many string metrics to measure the dissimilarity or mismatches between two different strings of uniform length. So It is basically determined by traversing both strings, character by character, and differing characters are counted. If the scenario arises where two strings are having a hamming distance of value zero, then both are called a 'perfect match'.

If length of two strings whose hamming distance is to be found out, is not equal, then we have used padding on smaller string to make it's length equal to larger string. Hamming distance of two strings (a,b) can be found out by following expression:

$$d(a,b) = \sum_{k=0}^{N-1} a_k \neq b_k \quad (1)$$

Where,
$d(a,b)$, is hamming distance between string $a$ & string $b$.
$a_k$, is the character of string $a$ at position $k$.
$b_k$, is the character of string $b$ at position $k$.

### c. Damerau-Levenshtein distance

Damerau–Levenshtein distance between two sentences is the least number of operations required to convert one sentence to another sentence. It is an extension to Levenshtein distance by including transpositions among its allowable operations along with three classical single character edit operations which includes insertions, deletion and substitution. Damerau–Levenshtein distance of two strings (a,b) can be found out recursively by following expression:

$$d_{a,b} = \begin{cases} \max(i,j) & \text{if } \max(i,j) = 0 \\ \min \begin{cases} d_{a,b}(i-1,j) + 1 \\ d_{a,b}(i,j-1) + 1 \\ d_{a,b}(i-1,j-1) + 1_{a_i \neq b_j} \\ d_{a,b}(i-2,j-2) + 1 \end{cases} & \text{if } i,j > 1 \text{ and } a_i = b_{j-1} \text{ and } a_{i-1} = b_j \\ \min \begin{cases} d_{a,b}(i-1,j) + 1 \\ d_{a,b}(j,i-1) + 1 \\ d_{a,b}(i-1,j-1) + 1_{a_i \neq b_j} \end{cases} & \text{otherwise} \end{cases} \quad (2)$$

Where,
$d_{a,b}$, is Damerau-Levenshtein distance between two strings $a$ and $b$.
$1_{a_i \neq b_j}$, is the indicator function which is 0 when $a_i = b_j$ and 1 otherwise.
Each recursive call matches one of the cases:
$d_{a,b}(i-1,j) + 1$, is deletion operation (from a to b)
$d_{a,b}(i,j-1) + 1$, is insertion operation (from a to b)
$d_{a,b}(i-1,j-1) + 1_{a_i \neq b_j}$ is match or mismatch depending on whether the respective characters are same.
$d_{a,b}(i-2,j-2) + 1$ is transposition operation between two successive characters.

## V. METHODOLOGY

The detailed working of our system is described in following sections:

### A. Hindi news content extraction and translation :

We have extracted Hindi news content along with its headlines from the Navbharat Times, as per the computed range of dates, according to the range of month and year, which are taken as an input. Then after preprocessing (i.e. removing irrelevant data from the text) the crawled Hindi news contents and headlines, these are then translated to English news headlines and English news contents using Google translator API. This translated English news contents acts as a baseline for the creation of parallel corpus.

### B. English news content extraction :

In this step, on the basis of each translated English news headline, the top ten links are extracted using Google search API. From these links, English news contents are extracted and then the English content having best token sort ratio i.e. number of similar words, with respect to the corresponding news content of baseline corpora is found out. Then the extracted English content is preprocessed.

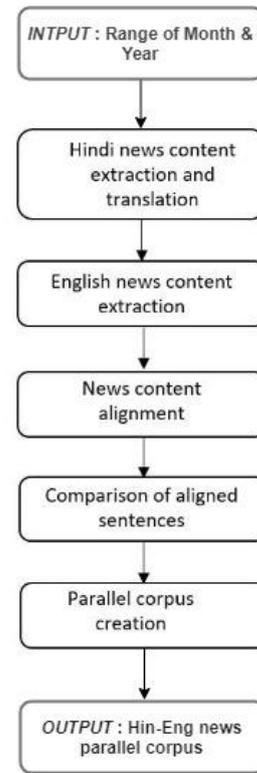

Fig. 1 : Methodology of automatic news parallel corpus creation

### C. News content alignment

The categorization of crawled Hindi news content, translated English news content and crawled English news content is done as per date, month and year and after that the respective documents are aligned based on bilingual correspondence in the subsequent stage of the process.

### D. Comparison of aligned sentences :

Firstly, the sentence boundary disambiguation of the Hindi news content, translated English news content and crawled English news content is done on the basis of a approach based on, end of a sentence delimiters like '!', '?', '.' for English and '!', '?', '।' for Hindi, simultaneously managing various exceptional cases by synthesizing a rule based system taking token length, abbreviations and decimal numbers into consideration.

Then each sentence in translated English news content is compared with all the sentences of crawled English content by finding its simple ratio of fuzzy string matching algorithm, which is based on levenshtein distance as described in following sub-section.

#### 1) Fuzzy String Matching algorithm:

Fuzzy string matching is a method based on conducting a human-like evaluation of the closeness or similarity of two phrases or words. In most of the cases, it requires identifying phrases or words which are the most close or similar to each other.

Here we are finding out its simple ratio, which involves a weighted comparison of the length of the two phrases, the changes required between each phrase, and whether each phrase could be found in the target entry. The distance metric used is levenshtein distance, which determines the changes or operations required to change one string to another. Levenshtein distance of two strings (a,b) can be found out by following expression :

$$lev_{a,b}(i,j) = \begin{cases} \max(i,j) & if \min(i,j) = 0, \\ \min \begin{cases} lev_{a,b}(i-1,j) + 1 \\ lev_{a,b}(i,j-1) + 1 \\ lev_{a,b}(i-1,j-1) + 1_{a_i \neq b_j} \end{cases} & otherwise \end{cases} \quad (3)$$

Where,
$1_{a_i \neq b_j}$ is the indicator function equal to 0 when $a_i = b_j$ and equal to 1 otherwise.
$lev_{a,b}(i,j)$, is the distance between first $i$ characters of $a$ and first $j$ characters of $b$.

Based on levenshtein distance metric, another metric 'value words' is used. It separates the string into a set of individual words, based on delimiters like dashes, spaces and anything else you'd like, and then each word is compared to each other word, summing up the smallest levenshtein distance linking any two words.

*E. Parallel corpus creation*

For the formation of a bi-lingual pair we have taken threshold values like 50%, 60%, 70% and 80%, which determines the percentage of similarity of crawled English news content to that of baseline as per the fuzzy string matching algorithm.

The sentences of crawled English news content having simple ratio more than that of the given threshold are extracted and the one having the highest simple ratio value is selected to be aligned with its corresponding sentence of Hindi news content. This bi-lingual pair are then saved as a part of parallel corpus for that particular threshold value.

The algorithm of the architecture of our prototype for Hindi-English news translation task is described below:

**Algorithm:** Algorithm for automatic parallel corpus creation
**Input:** Start = in(month & year), End = in(month & year)
**Output:** Collection of parallel sentences.
  *Initialisation:*
    func FuzzyStringMatching(parameters: Strings, Strings)
    var Date
    var Threshold
  *LOOP Process:*
  For Start Date ≤ Date ≤ End Date do
    1: Extract Hindi news content as well as Hindi news headlines date wise and preprocess them.
    2: Translate extracted Hindi news headlines into English news headlines.
    3: Crawl top 10 links by doing web search of the translated English news headlines and from these links extract English news content which is having best token sort ratio.
    4: Perform sentence alignment of the extracted Hindi news contents, translated English news content and extracted English news content
    5: While not EOF do:
         Similarity = call FuzzyStringMatching( Aligned translated English news contents, Aligned extracted English news contents )
         If Similarity ≥ Threshold
             Save corresponding sentence pairs of extracted Hindi news content and extracted English news content as a part of parallel corpus.
         Else
             Discard

VI. EXPERIMENTS, RESULTS AND DISCUSSIONS

For an input of 'December, 2017', 22,625 Hindi news were crawled from 1st December, 2017 to 31st December, 2017 i.e. in 31 days. And as we are crawling English news contents from the top 10 links from Google search, so along with that, the total English news crawled were 226,250 based on the translated English news headlines. The final output i.e. parallel corpora with respect to different threshold values based on the previously mentioned fuzzy string matching algorithm is described in Table I.

TABLE I
Number of sentences-pairs in parallel corpus for different thresholds

| | THRESHOLD | | | |
|---|---|---|---|---|
| | 50 | 60 | 70 | 80 |
| No. of sentence-pairs | 12,798 | 3,443 | 987 | 202 |

Some sample sentence pairs for all the threshold values taken into consideration, are mentioned as follows in table II to V:

TABLE II
Sample sentence pairs for threshold 50%

| Sr.No. | Threshold : 50% |
|---|---|
| 1 | यह स्कॉर्पिन श्रेणी की उन 6 पनडुब्बियों में से पहली पनडुब्बी है, जिसे भारतीय नौसेना में शामिल किया जाना है |
| | Named after the first Foxtort-class submarine, INS Kalvari was inducted into the Navy on December 8, 1967 |
| 2 | केंद्रीय सामाजिक न्याय एवं अधिकारिता राज्य मंत्री अठावले ने यहां जारी किए गए एक बयान में कहा, बाबासाहब अंबेडकर ने बौद्ध धर्म तब अपनाया जब उन्हें पूरी तरह भरोसा हो गया कि दलितों को हिंदू धर्म में न्याय नहीं मिलेगा |
| | Athawale, the minister of state for social justice and empowerment, said in a statement issued here |
| 3 | लाखों दलितों ने भी धर्म परिवर्तन किया |
| | Lakhs of Dalits converted as well |
| 4 | 63 वर्षीय जोंस की जीत से सीनेट में डेमोक्रैट पार्टी के सदस्य 49 हो गए हैं, जबकि रिपब्लिकन के 51 सदस्य हैं |
| | The victory of 63 year old gives the Democrats a clean sweep over Republican in statewide elections in 2017 |
| | इस जीत की अमेरिकी मीडिया में काफी चर्चा हो रही है |

| 5 | American media talks heavily about the resounding victory of the Democrats |
|---|---|

TABLE III
Sample sentence pairs for threshold 60%

| Sr.No. | Threshold : 60% |
|---|---|
| 1 | लाखों दलितों ने भी धर्म परिवर्तन किया |
| | Lakhs of Dalits converted religion as well |
| 2 | रिपब्लिकन पार्टी ऑफ इंडिया ए के प्रमुख अठावले ने बसपा प्रमुख मायावती को भी निशाने पर लेते हुए कहा कि वह बार बार धमकियां देने की जगह उन्हें बौद्ध धर्म अपना लेना चाहिए |
| | Athawale, the chief of the Republican Party of India (A), also slammed BSP chief Mayawati and said instead of issuing repeated threats, she should once and for all convert to Buddhism |
| 3 | यहां उन्होंने किसानों से संवाद करते हुए वादा किया था कि उनकी सरकार 'किसान मित्र' के रूप में काम करेगी |
| | Modi, who interacted with the farmers in Dabhadi on March 20, 2014, had promised that his government would work as a kisan mitra. |
| 4 | स्कूलों से उनके स्टाफ का वेरिफिकेशन कराने को भी कहा जा रहा है |
| | Schools must ensure police verification of both teaching and non-teaching staff |
| 5 | दूसरे चरण के लिए कुल 851 उम्मीदवार मैदान में हैं |
| | A total of 851 candidates are contesting in the second phase of Gujarat election for 93 seat |

TABLE IV
Sample sentence pairs for threshold 70%

| Sr.No. | Threshold : 70 % |
|---|---|
| 1 | दूसरे चरण के लिए कुल 851 उम्मीदवार मैदान में हैं |
| | A total of 851 candidates are in the fray for stage 2 |
| 2 | रिपब्लिकन पार्टी ऑफ इंडिया ए के प्रमुख अठावले ने बसपा प्रमुख मायावती को भी निशाने पर लेते हुए कहा कि वह बार बार धमकियां देने की जगह उन्हें बौद्ध धर्म अपना लेना चाहिए |
| | Athawale, the chief of the Republican Party of India (A), also slammed BSP chief Mayawati and said instead of issuing repeated threats, she should once and for all convert to Buddhism |
| 3 | छह और उत्पादक राज्य हैं जो एफआरपी का अनुपालन करते हैं |
| | There are six other producing states that follow FRP |
| 4 | उत्पादक क्षेत्रों से आपूर्ति में गिरावट और वायदा बाजार में मजबूती के रुख के कारण भी तेजी को समर्थन प्राप्त हुआ |
| | Further, fall in supplies from producing belts and firming trend in futures market also supported the upside |
| 5 | उन्होंने कहा कि संभवत: यह पहला नोटिस है जो किसी को सोशल मीडिया पोस्ट के लिए मिला है |
| | This was "probably the first notice" anyone had received for social media posts, she added |

TABLE V
Sample sentence pairs for threshold 80%

| Sr.No. | Threshold : 80 % |
|---|---|
| 1 | धनिया और हल्दी की कीमतें 100 - 100 रुपये की तेजी के साथ क्रमश: 6,100 - 13,100 रुपये और 8,700 - 11,900 रुपये प्रति क्विंटल पर बंद हुई |
| | Coriander and turmeric prices rose by Rs 100 each to conclude at Rs 6,100-13,100 and Rs 8,700-11,900 per quintal, respectively |
| 2 | उन्होंने यह कर लिया ! अनुष्का और विराट के लिए हमारी सबसे हार्दिक बधाई |
| | They made it ! Our heartiest congratulations to Anushka & Virat |
| 3 | उन्होंने बताया कि जीएसटी परिषद आगामी दिनों में बिजली, रियल स्टेट और पेट्रोलियम को जीएसटी के दायरे में लाने के मसले पर विचार कर रही है |
| | He said the GST Council was discussing the issue of bringing electricity, real estate, and petroleum under the GST ambit in coming days |
| 4 | माकपा की रीता मंडल वाम मोर्चा उम्मीदवार के तौर पर इस सीट के लिए चुनाव लड़ेंगी |
| | The CPM's Rita Mandal will contest as a Left Front candidate |
| 5 | सप्ताहांत की छुट्टी के बाद दोनों सदन की कार्यवाही सोमवार को फिर शुरू होगी जब चुनाव नतीजे घोषित किए जाएंगे |
| | After the weekend break, the two Houses will meet again on Monday when the poll results will be announced |

*a. System Configuration*

The system configuration used for this purpose is as follows:

TABLE VI
System configuration

| Processor | Intel(R) Core(TM) i7-6700 CPU @ 3.40GHz |
|---|---|
| RAM | 32 GB |
| System Type | 64-bit OS , x64-based processor |
| OS | Ubuntu 16.04 |

### b. Results and analysis

The complete result, where the average accuracy of the generated parallel corpus among different threshold values is compared with respect to different sentence matching metrics, is given in the table VII

TABLE VII
Accuracy of parallel corpora for different thresholds over metrics

| Threshold | Gestalt Pattern Matching | Hamming Distance | Demarau-Levenshtein Distance |
|---|---|---|---|
| 50 % | 49.79 | 71.73 | 33.09 |
| 60 % | 50.04 | 73.4 | 43.14 |
| 70 % | 66.6 | 77.98 | 56.3 |
| 80 % | 80.04 | 85.19 | 69.36 |

It can be observed that with increase in threshold values the accuracy of sentence matching metrics is also increasing. Among all the sentence matching metrics used, Hamming distance is giving better result compared to others for all threshold values shown in figure 2.

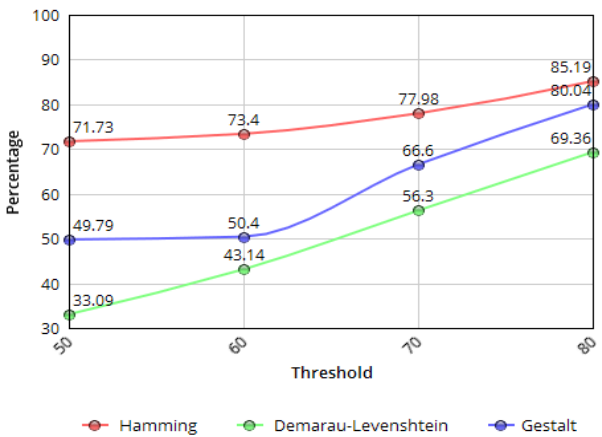

Fig. 2 : Linear plot of Accuracy of parallel corpora for different thresholds over metrics

### VII. CONCLUSION AND FUTURE SCOPE

We have conducted this experiment to create parallel corpus from the extracted comparable corpus for Hindi-English news translation task. And for the creation of corpus we have taken various threshold values for the extraction of bi-lingual sentence pairs from comparable corpora based on fuzzy string matching algorithm. Quality of the corpus is further tested with various performance metrics like Gestalt pattern matching, Hamming distance and Demarau-Levenshtein distance. Based on the results it can be observed that the accuracy of the parallel corpus is increasing with increase in threshold values.

We will try to increase the quantity of parallel corpus mainly for the threshold value of 80, as the similarity of bi-lingual sentence pair is highly noticeable compared to other threshold values. Further based on the parallel corpus generated, we will be trying to develop a Hindi-English news Machine Translation system.